\documentclass{article}

\usepackage{microtype}
\usepackage{graphicx}
\usepackage{subfigure}
\usepackage{booktabs} % for professional tables
\usepackage[table,dvipsnames]{xcolor}
\usepackage{epsfig}
\usepackage{pgfplotstable}
\usepackage{pgfplots}
\usepgfplotslibrary{groupplots}
\usepackage{amsmath}
\usepackage{amssymb}
\usepackage{bbm}
\usepackage{float}
\usepackage{booktabs}
\usepackage{multirow}
\usepackage{adjustbox}
\usepackage{verbatim}
\usepackage[T1]{fontenc}
\usepackage{graphicx}
\usepackage{caption}
\usepackage{paralist}%Compact lists 
\usepackage{siunitx}
\usepackage{nicefrac}
\usepackage{xspace}
\usepackage[colorinlistoftodos,textsize=footnotesize]{todonotes}

\usepackage[utf8]{inputenc}

\newcommand{\textcite}[1]{\citet{#1}}

\usepackage[autostyle, english=american]{csquotes}
\MakeOuterQuote{"}

\usepackage{url}

\usepackage{breakurl}

\usepackage{hyperref}

\usepackage[accepted]{icml2018}

\everypar{\looseness=-1 } %Please try and avoid dangling words on their own lines! 
\begin{document}

\twocolumn[
\icmltitle{What About Applied Fairness? }



\icmlsetsymbol{equal}{*}

\begin{icmlauthorlist}
  \icmlauthor{Jared Sylvester}{bah}
  \icmlauthor{Edward Raff}{bah,umbc}
\end{icmlauthorlist}

\icmlaffiliation{bah}{Booz Allen Hamilton}
\icmlaffiliation{umbc}{University of Maryland, Baltimore County}

\icmlcorrespondingauthor{Edward Raff}{raff\_edward@bah.com}

\icmlkeywords{Machine Learning, ICML}

\vskip 0.3in
]



\printAffiliationsAndNotice{}  % leave blank if no need to mention equal contribution

\begin{abstract}

Machine learning practitioners are often ambivalent about the ethical aspects of their products. 
We believe anything that gets us from that current state %--- in which we're too often entirely ambivalent about fairness ---
to one in which our systems are achieving some degree of fairness is an improvement that should be welcomed. This is true even when that progress does not get us 100\% of the way to the goal of "complete" fairness or perfectly align with our personal belief on which measure of fairness is used.
Some measure of fairness being built would still put us in a better position than the status quo.
Impediments to getting fairness and ethical concerns applied in real applications, whether they are  abstruse philosophical debates or technical overhead such as the introduction of ever more hyper-parameters, should be avoided. In this paper we further elaborate on our argument for this viewpoint and its importance. 
\end{abstract}

\section{Introduction}

General questions regarding the fairness of machine learning models have increased in their frequency and study in recent years. Such questions can quickly enter philosophical domains and subjective world views \cite{pmlr-v81-binns18a}, but are crucial as machine learning becomes integrated in the fabric of society. The attention and critical thought is well deserved as we see applications emerge which can dramatically impact people's lives and families, such as predictive policing \cite{pmlr-v81-ensign18a} and sentencing \cite{Chouldechova2017,pmlr-v81-barabas18a}.

Despite this, we argue that a significant potion of the machine learning community are missing important questions regarding how to maximize the amount of fair machine learning deployed in the world. In particular, there are practical considerations for applied fairness with respect to current fairness that are being ignored. Stated simply if we want to increase fairness of real world machine learning systems, we should not delay solutions over concerns of optimal fairness when there currently exists no fairness at all. 

As such we must ask: how do we maximize the number of people implementing/deploying fair/ethical machine learning solutions? We posit that the answer to such a question is to minimize the amount of \textit{mental} and \textit{computational} work that must be done to gain fairness. This applies to any practitioner with varying degrees of education and/or training in ethics and machine learning. If the incremental cost to deployment is too significant, we argue the concern of fairness will often be dropped in the name of expediency and financial cost. 

Under this general belief, we have identified three areas where we feel the community could increase social good by instead tempering its advance on some optimal notion of fairness. These areas relate to the debate around what ideal of fairness should be used (mental cost), over-reliance on trolley car hypotheticals (mental cost), and the nature of the algorithms people are developing (computational cost).

\section{The Unfair Criticism for the Wrong Fairness} \label{sec:which_fairness}

One of the largest impediments to stopping the adoption of fairness by non-expert practitioners is answering "\textit{what is fairness?}" There is a rich history of philosophical debate around this very question from which we can build upon as a community \cite{pmlr-v81-binns18a}. At the same time, a philosophical conclusion has not been reached after hundreds of years --- so we arguably should not expect there to be one true definition.
If people can not agree on a single definition of "fair" when defining it with natural language, why would we expect a single definition to be found when we move into the more rigorous, less ambiguous language of mathematics and algorithms?
Furthermore, the definition of "fair" may change over time as societal views change, unlike technical definitions we are used to working with. There may in fact be no one, final, universal definition of "fair" to be found.

Indeed many differing definitions and metrics for fairness and discrimination in predictive machine learning have been developed \cite{Romei2014,Kamiran2009,Hardt2016} and shown to be at some level incompatible with each other \cite{Hardt2016}. This is focused on primarily binary prediction problems in ML; let alone nascent definitions in sub-areas like recommender systems~\cite{pmlr-v81-burke18a,pmlr-v81-ekstrand18b}, regression~\cite{Calders2013,Berk2017}, and clustering~\cite{Chierichetti2017} or those that may have been defined in neighboring fields like economics~\cite{10.2307/1810007}. 

Given these competing definitions of fairness, it is important that we as a community avoid being overly critical on what \textit{specific} definition of fairness is selected for an individual project or system. For those applications where no measures of fairness are currently considered, we should even go further and applaud and encourage the selection of \textit{any} reasonable fairness criteria, even if it is not the one we would have personally preferred. 

Doing so immediately increases the amount of fairness, by some metric, in the deployed world --- which we argue is intrinsically of greater social good than leaving the question of fairness wholly unaddressed. The implementers of the system, by selecting any measure, are now invested in the fairness of their product and thus may become more open to improving the fairness as a type of \textit{feature}.  Even if another measure is objectively superior given some context, having a less-ideal metric implemented opens the door for revisiting and adjusting the fairness portion of the system at another point in time.

In addition, a machine learning system of suboptimal fairness may still be more fair than the non-quantitative, human system it augments or replaces. An only-partially fair quantitative system may be preferred because it can be measured, logged and inspected in a way that no human-driven, qualitative decision making can be. This greater legibility can lead to greater transparency.

Encouraging this could prove to be an advantageous path of least resistance. Not only does it allow for a transitional nature, but can yield positive network effects within larger organizations. For example, Team A gets to add a monthly update report that fairness as added as a feature, which could get other mangers or engineers thinking about fairness for their project. This may not happen if members of the team fear being censured for choosing the ``wrong'' metric of fairness, or for implementing a system which increases fairness without completely maximizing it on some measure.

An important component of this success is respectful discourse between groups on disagreements about what is or is not fair, and openness about how one is measuring fairness in a given system. If these do not exist, disagreements may devolve to stronger accusations and acrimony. 

\textcite{Johndrow2017} highlighted an example of this with the maligned \textsc{compas} system for predicting criminal recidivism. \textcite{Angwin2016} from ProPublica published an article about bias in the \textsc{compas} system. In response, the company which developed \textsc{compas}, Northpointe, released a report showing the metrics by which their system was fair~\cite{Dieterich2016}.
Clearly the issue under consideration is of critical importance, to a degree such that the debate about what the best measure of fairness is and how to make the system more fair as a whole should be mandatory and continuous. But the nature of how this debate has unfolded (in this particular instance) has lead to considerable negative publicity when it appears that Nortpointe made an earnest good faith effort to address the issue before it became newsworthy.
The issue appears not to be fair-vs-unfair, but which of two competing and somewhat incompatible definitions of fairness should be prioritized.

As a community we must avoid exchanges like \textsc{compas} to avoid scaring off future leaders and decision makes from the issue of fairness (and machine learning more broadly). Put simply, \textsc{compas} sets a precedent for social risk via negative publicity even when attempting to imbue machine learning systems with fairness. Even if one were to go well beyond what Northpointe did, there is still a risk of censure from critiques simply because they may adopt a different definition of fairness. This risk may prevent adoption, and thus lower the total fairness within the world. 

If we instead accept that there is no single supreme definition of fairness, the situation can be improved. When we accept that others may not have considered certain factors in selecting their fairness measure, or may have reached their conclusion under different but equally valid philosophical beliefs, the conversation about fairness can be lifted to a more civil and less accusatory tone. In doing so the social risk can be transformed into social reward, as feedback will no longer be perceived as an attack that must be defended --- but as genuine interest from the larger community. 

\section{Should Autonomous Vehicles Brake for the Trolley Car?}

The trolley car problem \cite{foot1967problem,thomson1976killing} has been the subject of much debate recently, coinciding with the increased interest in both fairness and autonomous vehicles. While many variants exist, the general trolley car problem is as follows: 
if a vehicle continues on it's current path, it will kill five people in its way; if some action is taken it can instead strike and kill only a single person.
The specifics of the dilemma change (if the vehicle continues driving it will hit a child; if it swerves it will kill its passenger), but at its core it is a contrived situation with a set of exclusively bad outcomes.

With self-driving cars on the precipice of deployment, the trolley car problem makes intuitive sense for study. Hardware failures, sudden changes in environment (like an earthquake), or actions of bystanders/non-autonomous vehicles are all factors outside of a self-driving agent's control that could lead to a potentially fatal situation. All of this is made more pertinent due to the first, unfortunate, death at the hand of an autonomous vehicle \cite{Lee2018}. 

Even before this sad death, many have been debating the trolley car problem and arguing that a solution is needed for deployment \cite{Achenbach2015,Corfield2017,Lin2016,Goodall2016}. This circles back to the problems we discussed in \autoref{sec:which_fairness} on what measure of ethical behavior we should be using to decide who lives and who dies in the myriad of possible trolley car scenarios? Surveys  reveal that people prefer that cars be willing to sacrifice the driver, but simultaneously would not personally want to own such a car \cite{Bonnefon2016}.  That this would create an  dichotomy is understandable, but it makes reaching a consensus on what should be done difficult. Further studies have looked at presenting varied trolley car scenarios and simply asking people which way the car should swerve, and then attempting to quantify the resulting empirical ethos~\cite{Shariff2017}. 

Despite all of these questions of research and debate, we do not see it asked: do drivers today consider the trolley car problem when they are about to enter an accident? We argue that no such consideration exists today or even could with human drivers. The small amount of time to react in any such scenario likely means people are simply relying on gut reactions and are not performing any meaningful consideration of who will and will not survive an accident. Nor do we prepare people to make these sorts of decisions: no ethical training or testing is undertaken before issuing people with drivers' licenses.

If people are not considering this problem today, why should we \textit{require} self driving cars to do the same? It results in a moving of goal posts, requiring cars to reach super-human abilities before we let them take over a task.%
\footnote{Some argue that AI should only have to be as ethical as the humans whose decisions they are supplanting. Other claim that since AI may have super-human abilities, it is not unreasonable that they have super-human ethical responsibilities. We would contend that holding AI to a higher standard than humans may be acceptable, but holding them to a standard of \textit{perfection} is not.}
If self driving cars can reduce the number of fatalities by 90\% \cite{Bertoncello2015}, then we reduce the incident rate of trolley car situations by 90\%. In this way we are in a sense solving the trolley car problem by reducing its frequency, as the best possible scenario is the one where the trolley car problem never occurs.  We argue this increases social good without having to solve such a difficult problem, and that delaying deployment until such a satisfactory solution is obtained may in-fact needlessly delay improved safety for everyone. 

We take a moment to emphasize that we are not arguing self driving cars should be deployed as soon as possible. Considerable and thorough safety and validation testing should be mandatory before public deployment; corners can not afford to be cut. We are  arguing that certain fairness considerations that are being debated, such as the trolley car problem, have been imbued with an importance beyond the reality of their application. 

Along these lines, we need to further consider what situations will lead to trolley car problems. It seems likely that one of the most likely culprits is mechanical failures: breaks stop working effectively, steering or sensor systems may malfunction, etc. In such a case, even if the car had an oracle that solved the trolley car problem, it is not obvious to us that it would be able to execute on that solution due to the aforementioned mechanical failure. 

Going further, even if we did have a oracle that can solve the trolley car problem, we likely could not effectively use it. This is because the car itself will need to be predicting people, their ages, the risk of fatality, an a myriad of other factors that would be necessary inputs to the trolley car problem. But each of these predictions will have their own error rates, and some, like risk of fatality, may not even have any reliable models developed. Realistically any trolley car  solution would also require an understanding of risk and uncertainty about the situation itself. This is an issue we don't see discussed, and is contributes to why we feel a trolley car solution is an unreasonable expectation. 

To delay a potential life-saving innovation is itself deadly. We are engaging in a real-life meta-trolley problem:
our meta-trolley is currently running on a track that allows human drivers to kill a million people a year~\cite{who2015road}, and could be switched to an alternate track that may be far less deadly. Meanwhile we stand by arguing about the propriety of pulling the lever to allow the meta-trolley to switch tracks.

\section{Fairly Complicated Fair Algorithms}

We've discussed two situations in which the emphasis on getting fairness exactly right may lead to reduced fairness in practice. Now we discuss a matter with regard to practitioners in making fairness algorithms as usable as possible. This means reducing the number of hyper parameters, and computational and cognitive costs in adding fairness to current algorithms, an issue we feel is under studied. 

A common issue is the introduction of multiple new hyper parameters to an algorithm, in addition to the ones that existed before \cite{pmlr-v28-zemel13,Louizos2016,Edwards2016}. This can get particularly out of hand when multiple different parts of the model must be specified for any new problem \cite{Johndrow2017}. Such solutions necessitate a more expensive parameter search, thus increasing the financial cost of developing deployable solutions. This reduces the incentive for companies to invest in the time to make fair models, and thus should be something we try to  minimize.

While we have no expectation of a magic black box which will produce fair algorithms and require no work, we do believe there is room for considerable simplification of the approaches being developed. Having one or zero hyper parameters may not lead to a perfectly optimized balance between fairness and predictive performance, but it may lead to faster adoption and integration within organizations today, thus increasing fairness from our current baseline.

In a similar vein, we would like to see research along automatically selecting a measure of fairness to optimize for and providing human readable reports about what the ramifications would be. As far as the authors are aware, these two notions have yet to receive study in the machine learning community. 
The automatic selection of a fairness metric could be done with respect to a maximum acceptable loss in accuracy (e.g., which measure can be maximally satisfied at a fixed cost?). Though the solution may not be optimal, it could prove better than the default state of no fairness consideration.

A tool that can generate human readable reports on the impacts of different fairnesses measures and provide some "map" of the potential options would also provide value. It better enables product developers and practitioners who are not experts to weigh the costs of fairness and potentially integrate them, as well as the impacts of any measure selected in the aforementioned auto-fairness idea. 

The goal of all of these preferences for usable fair algorithms is not to directly solve fairness by any means; but to maximize social good in the near and long term. They create a path of least resistance for novices who are concerned about fairness so that \textit{something} can be integrated immediately.  This also opens the door to future exploration and improvement of fairness as its own feature, and provides, in our opinion, a viable method for integration into the maximal number of systems. If such work continues to be unstudied, we may leave businesses and developers a daunting task: a whole world of literature, competing definitions, and philosophical questions fraught with ethical and social complexities that must be understood before even being able to start. The apparent gap itself may become the biggest deterrent to adoption, and so we wish to implore the community to build these bridges. 

\section{Conclusions} \label{sec:conclusion}

Our current machine learning systems are becoming more powerful and being deployed more widely each day, and yet they --- and their creators --- are often completely oblivious to issues of fairness. There is a broad chasm between the current state of machine learning and ideally ethical systems. It is our contention that we should welcome any efforts which narrow that gap, even if they fall short of bridging it completely. 

We believe that some fairness is better than no fairness. Arguments, attitudes and techniques for perfect fairness are retarding out ability to get any improvements relative to the status quo. We should not let the perfect be the enemy of the good.

We call on people in this discussion to realize that other researchers and practitioners are trying to make the world better and more just, even if they aren't making the exact improvement that you might prefer. We do not mean that anyone should be beyond reproach, merely that we should aim to make critiques constructively and civilly so that we can work together toward a more fair society.

We propose that researchers and practitioners in this field should not ask  ``does this meet some Platonic ideal of fairness?'' but rather they should be concerned with ``does this increase the amount of fairness in the world?''
% \todo{Compare to No Free Lunch? We know better than to ask for a classifier with ideal accuracy and are content with better accuracy than we previously had. Why not the same for fairness?}

% \Urlmuskip=0mu plus 1mu\relax
\bibliographystyle{icml2018}
\small
\bibliography{Mendeley,refs}

\end{document}